\documentclass[10pt,twocolumn,letterpaper]{article}

\usepackage{cvpr}
\usepackage{times}
\usepackage{epsfig}
\usepackage{graphicx}
\usepackage{amsmath}
\usepackage{amssymb}
\usepackage{algorithm}
\usepackage[noend]{algpseudocode}
\usepackage[pagebackref=true,breaklinks=true,letterpaper=true,colorlinks,bookmarks=false]{hyperref}

 \cvprfinalcopy % *** Uncomment this line for the final submission

\ifcvprfinal\fi
\begin{document}

%%%%%%%%% TITLE
\title{A Generative Model For Zero Shot Learning \\ Using Conditional Variational Autoencoders}

\author{ Ashish Mishra$^{1}$ , Shiva Krishna Reddy$^{1}$, Anurag Mittal, and Hema A Murthy\\
{ Indian Institute of Technology Madras }\\
{\tt\small \{mishra,shiva,amittal,hema\}@cse.iitm.ac.in}
}

\maketitle
%\thispagestyle{empty}

%%%%%%%%% ABSTRACT
\begin{abstract}
 Zero shot learning in Image Classification refers to the setting where images from some novel classes are absent in the training data but other information such as natural language descriptions or attribute vectors of the classes are available. This setting is important in the real world since one may not be able to obtain images of all the possible classes at training. While previous approaches have tried to model the relationship between the class attribute space and the image space via some kind of a transfer function in order to model the image space correspondingly to an unseen class, we take a different approach and try to generate the samples from the given attributes, using a conditional variational autoencoder, and use the generated samples for classification of the unseen classes. By extensive testing on four benchmark datasets, we show that our model outperforms the state of the art, particularly in the more realistic generalized setting, where the training classes can also appear at the test time along with the novel classes. 
\end{abstract}

%%%%%%%%% BODY TEXT
\section{Introduction}

\noindent 
Availability of labeled image data has helped in making great advances in Computer Vision. However, even the largest image dataset i.e Imagenet \cite{imagenet_cvpr09} has only 21841 classes, with many classes having very few images. Thus collecting and training on sufficient number of images from all the classes of images that may occur in practice is a very difficult task. Moreover, new classes come into existence every day and images of certain classes may be rare and difficult to obtain. Human beings are excellent at recognizing novel objects that have not been visually encountered before. For instance given the information that an \textit{auroch} is \textit{an ancient cow, has large horns, has large build}, one can easily identify an image of an auroch from other animals such as a pig or sheep although one hasn't seen an auroch before. Zero shot learning tries to capture this intuition by assuming that some other information about the novel class is available although no image from that class is available in the training set \cite{conf/cvpr/HuangEEY15,conf/aaai/LarochelleEB08,journals/pami/LampertNH14,conf/cvpr/RohrbachSS11,conf/eccv/YuA10}. This extra information is typically in the form of attributes or textual descriptions.

More formally, let $X_{tr}$ and $Y_{tr}$ represent the training images and their class labels respectively. Similarly, let $X_{te}$ and $Y_{te}$ represent the test images and their corresponding labels. The zero shot setting states that $Y_{te}\not\subset Y_{tr}$. However for each label $y_i$ in $Y=Y_{tr}\cup Y_{te}$, we have an embedding vector, called class embedding vector $A_i$, that is semantically related to the class corresponding to that label. This vector could come from other modalities, such as language and may be obtained using different approaches such as manually or automatically annotated attributes (for e.g \textit{word2vec}). Recently zero shot learning has emerged as an active area of research in the interplay between vision and language \cite{CONSE,DEVISE,labelEmbeddingImgClassification,SJE,ESZSL2015,SYNC2016,SAE2017}. It is an interesting area of research since the models for this problem can help understand how well language concepts translate to visual information.

\begin{figure}
\caption{Illustration of ZSL}

\begin{centering}
\includegraphics[scale=0.35]{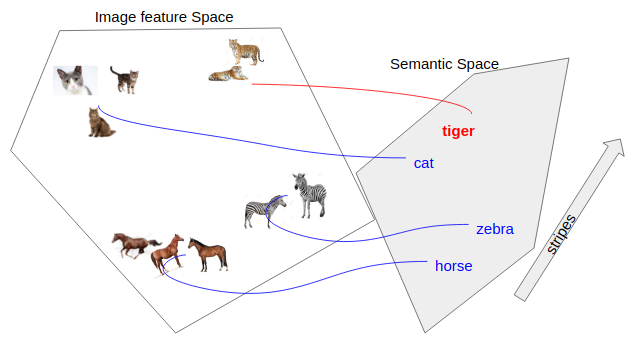}
\par\end{centering}
Suppose that \textit{cat}, \textit{horse} and \textit{zebra} are the training classes. The aim of zero shot learning is to correctly classify a new class image such as \textit{tiger} by relating it to them images of seen classes. The relationship between the classes themselves can be provided by the class embedding vectors but needs to be translated to the image domain. 
\end{figure}

If the classes are modeled accurately via the embedding vectors, the problem can be viewed as finding a relation between the embedding vector of a class and the visual features of the images in that class. Most zero shot learning approaches learn a projection from image space to the class embedding space via a transfer function. For a novel class image given at test time, the class with the closest class embedding vector to the projection in the class embedding space is assigned. Similarly it is also possible to learn a mapping function from class embedding to the image space. In the first, the mapping is a simple linear function with various kinds of regularizations. The challenges faced in learning such a mapping are well documented, the primary issue being that of domain shift first identified by \cite{TransductiveZSL}. The mapping learned from the seen class images may not correctly capture the relationship for unseen classes as the space may not be as continuous and smooth in the image domain. The image space may also be more complicated than the semantic space due to the complex image generation process.
Recent advances in unsupervised learning have led to better architectures for modeling the statistical image generation process, primary among them being generative adversarial networks (GANs) \cite{GAN14} and variational autoencoders (VAEs) \cite{VAE13}. These models can also be used for condition specific image generation \cite{CVAE_2015}. For example one can generate images conditioned on attributes. A natural question to ask is: How much can this attribute conditional image generation, generalize to unseen classes and can this be used for zero shot learning? 
 
In this work, instead of directly modeling a transfer function, we view problem as a case of missing data, and try to model the statistical image generation process via VAEs conditioned on the class embedding vector. The missing data for the unseen classes is filled using such generated image data.

ZSL models are typically evaluated in two ways. In the standard setting \cite{ZSL2009}, it is assumed that the train and test classes are disjoint $\left(Y_{tr}\cap Y_{te}=\Phi \right)$ i.e the training class images do not occur at test time. However, this is hardly true in the real world and can artificially boost results for the unseen classes without regard to the train class performance. Hence the Generalized zero shot setting has been proposed \cite{GenZSL2016} where both train and test classes may occur during the test time (Note that the latter setting is much harder than the former since the classifiers are typically biased either towards the classes seen at training time or those unseen at training time). We present our evaluation on both settings but obtained the greatest improvements in the much harder generalized setting. We follow the evaluation protocol recently proposed by \cite{xianCVPR17} that ensures that the models are not pre-trained on any of the test classes. The main contributions of this paper are as follows:

\begin{itemize}
\item We present a different approach to the zero shot problem by viewing it as a missing data problem. We train a Conditional Variational Autoencoder to learn the underlying probability distribution of the image features$(X)$ conditioned on the class embedding vector$(A)$. We show that such an approach reduces the domain shift problem that is inherent in the methods that learn a simple mapping as the image generation process is modeled in a much more sophisticated way.
\item By extensive testing on four benchmark datasets, we show that out model provides significant improvement over the state-of-the-art, particularly in the much harder generalized zero shot setting.
\item Since the model is able to generate image features of previously unseen classes, the results obtained in this work are promising and do provide evidence that conditional variational autoencoders indeed capture the underlying image generation process.  It also vindicates our conjecture about the generalisability of generative models for new tasks.
\end{itemize}

The paper is organized as follows: In section-2 we present a brief overview of the various existing methods for zero shot learning. Section-3 presents the motivation, and a description of the proposed approach. Section-4 describes the evaluation settings. Section-5 presents the experiments, results, and a comparison with existing methods. Finally in section-6, we present many directions for future work and a summary of our contributions. 

\section{Related Work}
\begin{figure*}
\caption{Network Architecture}

\begin{centering}
\includegraphics[scale=0.5]{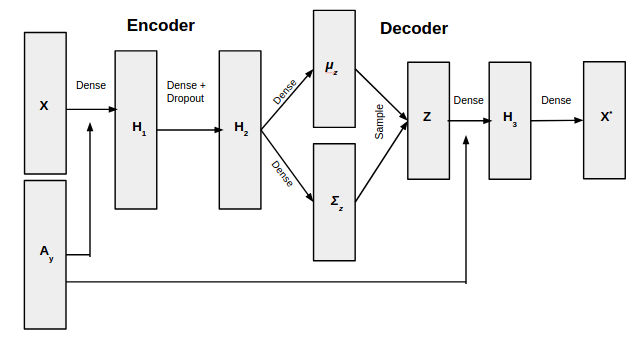}
\par\end{centering}
The input $x$ and the semantic class embedding vector $A_y$ are concatenated and passed through a dense layer, followed by dropout and another dense layer. This is followed by a dense layer, that outputs $\mu_z$ and $\Sigma_z$. A $z$ is sampled from the variational distribution $\mathcal{N}(\mu_z,\Sigma_z)$. The sampled $z$ is projected to a hidden layer via dense connections and then to the image space to reconstruct the original $x$. All activations are ReLU except the outputs of encoder and decoder which are linear. 
\label{net}
\end{figure*}
Zero shot learning was first introduced by \cite{ZSL2009}, where they consider disjoint train and test classes and propose an attribute based classification. Other traditional methods are based on learning an embedding from the visual space to the semantic space. During test time, for an unseen class example, the semantic vector is predicted and the nearest neighbor class is assigned \cite{wang2016relational,socher2013zero,verma2017simple}. The embedding is learned via a parameterized mapping. The most popular approach to ZSL is learning a linear compatibility between the visual and semantic space  \cite{labelEmbeddingImgClassification,DEVISE,SJE}. \cite{ESZSL2015,SAE2017} provide novel regularizations while learning a linear compatibility function.  ESZSL \cite{ESZSL2015} models the relationship between image features and attributes, via a simple compatibility function while explicitly regularizing the objective. SAE \cite{SAE2017} adds an autoencoder loss to the projection which encourages reconstructability from attribute space to the visual space. 

Image classification problems generally result in nonlinear decision boundaries. Linear methods have strong bias and often are not sufficient to model the problem. Hence, non linear compatibility learning methods have also been proposed. LATEM \cite{xian2016latent} proposes piecewise linear multimodal learning which learns nonlinear compatibility function and CMT \cite{socher2013zero} trains a neural network with one hidden layer with tanh activations and uses novelty detection for discriminating between seen and unseen classes in the generalized setting.

In another popular approach to ZSL, the seen class attributes are treated as the basis vectors \cite{CONSE} which map images(visual features) into the semantic embedding space via the convex combination of the class label embedding vectors weighed by the predictive probabilities for the different training class labels.  SYNC \cite{SYNC2016} tries to align the semantic space to the image space by learning manifold embedding of graphs composed of object classes. \cite{kodirov2015unsupervised} present a sparse coding framework based on an unsupervised domain adaptation for ZSL. 

A key component of zero-shot learning is the semantic embedding of the class labels. Previous work in ZSL has used human labeled visual attributes \cite{farhadi2009describing} to help detect unseen object categories for the datasets where such human labeled attributes are available. This work also uses the attribute vectors as the semantic class embeddings. Distributed representations of the class name such as \textit{word2vec} \cite{word2vec} have also been used as the semantic embedding. In this work we use word2vec for datasets where attribute vectors are not available. 

Prior work on conditional image generation based on textual descriptions was successful in generating synthetic images that appear natural \cite{Text2Img}. The authors also generate real looking images in the zero shot setting. This serves as the primary motivation behind our approach. However, we work in the feature space instead of the image space.

\section{Method Description}

We are given a set of train classes (also called seen classes) $\mathcal{Y}_{s}=\{y_{s}^{1},y_{s}^{2},...y_{s}^{n}\}$ and
a set of test classes (also called unseen classes) $\mathcal{Y}_{u}=\{y_{u}^{1},y_{u}^{2},...y_{u}^{m}\}$. 
 The zero shot setting states that $\mathcal{Y}_{u} \not \subseteq \mathcal{Y}_{s}$.
For each class $y$ in $\mathcal{Y}=\mathcal{Y}_{u}\cup\mathcal{Y}_{s}$,
we have a class semantic embedding vector $A_{y}$, that $\textit{describes}$
the class. We are given $d$-dimensional labelled training data from
the seen classes $\mathcal{Y}_{s}$ , i.e $\{X_{s},Y_{s}\}$. The
goal is to construct a model $f\,:\,\mathbb{R}^{d}\rightarrow\mathcal{Y}_{u}$,
that can classify the examples from the unseen classes $\mathcal{Y}_{u}$.
In the generalized zero shot setting, we aim to construct a more generic model
$f_{gen}\,:\,\mathbb{R}^{d}\rightarrow\mathcal{Y}_{s}\cup\mathcal{Y}_{u}$,
that can classify the data from both the seen and unseen classes correctly. The latter setting is much more difficult than the former for two reasons. Firstly, there are more classes in the second setting which leads to more confusion. Secondly, the seen and unseen classes may come from different probability distributions, which could degrade the performance on either the seen or the unseen classes. 
 
The Variational Autoencoder \cite{VAE13} is a graphical model, that tries to relate the distribution of the hidden latent representations $z$ to that of the data $x$. In variational inference, the posterior $p(z|x)$ is approximated by a parametrized distribution $q_{\Phi}(z|x)$    called the variational distribution. The lower bound for $p(x)$ can be written as follows: \\
\[
\mathcal{L}(\Phi,\theta;x)=-KL(q_{\Phi}(z|x)||p_{\theta}(z))+\mathbb{E}_{q_{\Phi}(z|x)}\left[\log p_{\theta}(x|z)\right]
\]
\\
\noindent  The network has
two components, an encoder network $E$ with parameters $\Phi$
and a decoder $D$ with parameters $\theta$.  Here the $p_{\theta}(x|z)$ can be seen as a decoder, from
latent space to data space, while $q_{\Phi}(z|x)$ can be seen as
an encoder from data space to latent space. Note that one can also view this
optimization as minimizing the reconstruction loss with the KL divergence
as the regularizer. Conditional Variational Autoencoders (CVAE) first introduced by \cite{CVAE_2015} maximize the variational
lower bound of the conditional likelihood $p(x|c)$ in a similar manner, which helps to
generate samples having certain desired properties (as encoded by $c$). 

\[
\mathcal{L}(\Phi,\theta;x,c)=-KL(q_{\Phi}(z|x,c)||p_{\theta}(z|c))
\]
\[
+\mathbb{E}_{q_{\Phi}(z|c)}\left[\log p_{\theta}(x|z,c)\right]
\]

In this work, we train the conditional variational autoencoder to generate
the data features $x$, given the conditional variable $A_{y}$ (the semantic embedding vector of a particular class). This helps us to model $p(x|A_y)$.

\subsection{Encoder}

The encoder generates the probability distribution $q(z|x,A_{y})$
 which is assumed to be an isotropic gaussian. $q$ is a distribution
over the latent space that gives high probability mass to those $z$
that are most likely to produce $x$ (which belongs to class $y$).
Thus, the encoder takes in a data feature point $x$ concatenated with
the semantic embedding $A_{y}$ and outputs the parameter vector of the Gaussian, i.e $\left(\mu_{x},\Sigma_{x}\right)$.

\subsection{Decoder}

The decoder tries to map the latent space to the
data space. For an input $\{z\circ A_{y}\}$ it tries to reconstruct that
$x$ of class $y$ which is most likely under the latent variable
$z$. If the CVAE is properly trained, one can use the decoder part
of the network to generate any number of samples of a particular class
using a simple algorithm : Sample $z$ from a standard normal, concatenate
$A_{y}$, and pass it through the decoder. 
\\

We model both the encoder and the decoder using neural networks. We observed that training with two hidden layers in the decoder quickly overfits to the seen classes, even with batch normalization and dropout. Challenges in training deep VAEs have been well documented \cite{conf/nips/SonderbyRMSW16}.  

During training, for each training datapoint $x^{(i)}$, we estimate
the $q(z^{(i)}|x^{(i)},A_{y_{i}})=\mathcal{N}(\mu_{x_{i}},\Sigma_{x_{i}})$
using the encoder. Then, a $\widetilde{z}$ is sampled from $\mathcal{N}(\mu_{x_{i}},\Sigma_{x_{i}}).$
We pass the $\widetilde{z}$ concatenated with $A_{y}$ to the decoder
and expect it to reconstruct $x$. We also want the $q(z)$
to be close to the standard normal distribution and include the KL divergence term in the optimization. Let $x$ be the input
to the encoder and $\hat{x}$ be the reconstructed output, the training loss becomes: 

\[
\mathcal{L}(\theta,\Phi;x,A_{y})=\mathcal{L}_{reconstr}(x,\hat{x})+KL\left(\mathcal{N}(\mu_{x},\Sigma_{x}),\mathcal{N}(0,I)\right)
\]
\noindent
We use $L_{2}$ norm for the reconstruction loss. The KL divergence
term has a nice closed form expression (see \cite{VAE13}) when the distributions are Gausian. 
The network is shown in the Figure-\ref{net}. Once the network is trained, one can sample any number of examples from each unseen class, since their semantic class embedding vectors are known. We call this the \textit{pseudo train data}. Once the data is generated, one can train any classifier for the unseen classes. We use an SVM classifier \cite{SVM} in this work. Note that the CVAE can possibly be replaced with any other generative model such as GAN \cite{GAN14} or it's many variants  as well.  The pipeline of the zero shot classification algorithm (Algorithm-1) is now simple:

\begin{enumerate}
\item Using $X_{train},Y_{train},A$ , train the Conditional Variational
Autoencoder. 
\item For each unseen class $y_{u}^{(i)}$ , generate samples of datapoints
belonging to that class. For this, sample $N$ latent vectors ($z$)
from a standard normal, concatenate them with $A_{y_{u}^{(i)}}$
and use it as the input to the decoder. We call generated output the pseudo traindata.
\item Train an SVM classifier using the pseudo traindata. 
\end{enumerate}
One straightforward way to extend our model to the generalized setting is to train the SVM with both the original training data of the seen classes, and the generated data of the unseen classes. However, we noticed that this leads to a bias towards the seen classes during classification.  Hence, we also generate pseudo data for seen classes along with unseen classes in the generalized zero shot setting. The change comes in step-6 of the algorithm, where $\mathcal{Y}_u$ is replaced with $\mathcal{Y}_s\cup \mathcal{Y}_u$. Thus, our model provides an easy extension to the much harder generalized zero shot setting.
\begin{algorithm}
\caption{CVAE-ZSL}\label{alg:euclid}
\begin{algorithmic}[1]
\Procedure{ZSL Classifier}{$X_s,Y_s$}
\State $N=300$ // Chosen via cross-validation
\State $model$ $\gets$ Initialize $Encoder$, $Decoder$
\State Train $model$ on $X_s,Y_s$
\State \textit{S}= $\Phi$
\For {$y_u\in \mathcal{Y}_u$ }
\For {$i$ in $[1,2,..N]$}
\State $z \sim \mathcal{N}(0,I)$ %\Comment{Sample from prior}
\State $V_i=y_u\circ z$
\State $X_i\gets Decoder(V_i)$ %\Comment{Sample from estimated distribution}
\State \textit{S}$\leftarrow$\textit{S} $\cup \{(X_i,y_u)\}$
\EndFor
\EndFor
\State $clf\gets SVM$
\State fit $clf$ on \textit{S}
\State \textbf{return} $clf$ 

\EndProcedure
\end{algorithmic}
\end{algorithm}

\section{Evaluation Protocol}
Typically, zero shot learning methods are evaluated in two settings. In the first setting, we make an assumption that seen classes do not occur at test time \cite{ZSL2009}, i.e $Y_{tr}\cap Y_{te}=\Phi$. We call this the disjoint assumption. However this is a very strong assumption and is usually not true in the real world. Nonetheless such a setting is useful to evaluate how the training generalizes to the unseen classes. In the generalized zero shot setting \cite{GenZSL2016}, there are no such assumptions made on the test data. Both seen and unseen class' images can occur at test time. Such a setting while being realistic, is also very difficult compared to the disjoint assumption. We follow the benchmark laid out recently by \cite{xianCVPR17,XLSA17} which ensures that our feature extraction model is not pretrained on any of the test classes. 

\section{Experiments}
We present our results on four benchmark datasets: Animals with Attributes (AwA) \cite{journals/pami/LampertNH14,XLSA17}, CUB-200-2011 Bird (CUB)\cite{CUB2010}, SUN Attribute (SUN) \cite{SUN2012} and Imagenet \cite{imagenet_cvpr09}. We also present results on AwA, CUB, and SUN for generalized zero shot setting. While AwA is a medium sized coarse grained dataset, SUN and CUB are medium sized fine grained datasets. We also present our results on the large scale Imagenet dataset. The 1000 classes of ILSVRC2012 \cite{ILSVRC15} are used as seen classes, whereas 360 non-overlapping classes of ILSVRC2010 are used for testing (same as \cite{SSVOC2016,conf/cvpr/FuXKG15}). The details of test train splits in each dataset are presented in Table-\ref{tab1}. We use keras \cite{chollet2015keras} with tensorflow backend \cite{tensorflow2015-whitepaper} for the implementation. The code will be made publicly\footnote{Github link with data and code will be posted} available to enable reproducibility.

\begin{table}
\caption{Dataset Details}
\begin{centering}
\begin{tabular}{|c|c|c|}
\hline 
\textbf{Dataset} & \textbf{\#images} & \textbf{seen/unseen}\tabularnewline
\hline 
\hline 
AWA-1 & 30475 & 40/10\tabularnewline
\hline 
AWA-2 & 37322 & 40/10\tabularnewline
\hline 
CUB & 11788 & 150/50\tabularnewline
\hline 
SUN & 14340 & 645/72\tabularnewline
\hline 
Imagenet & 254000 & 1000/360\tabularnewline
\hline 
\end{tabular}
\par\end{centering}
\label{tab1}
\end{table}

\begin{table*}
\centering{}%
\begin{tabular}{|c|c|c|c|c|}
\hline 
\textbf{Method} & \textbf{AWA-1} & \textbf{AWA-2} & \textbf{CUB} & \textbf{SUN}\tabularnewline
\hline 
\hline 
DAP \cite{lampert2014attribute} & 44.1 & 46.1 & 40.0 & 39.9\tabularnewline
IAP \cite{lampert2014attribute} & 35.9 & 35.9 & 24.0 & 19.4\tabularnewline
CONSE \cite{CONSE} & 45.6 & 44.5 & 34.3 & 38.8\tabularnewline
DeViSE \cite{DEVISE} & 54.2 & 59.7 & 52.0 & 56.5\tabularnewline
ALE \cite{akata2016label} & 59.9 & 62.5 & 54.9 & 58.1\tabularnewline
SJE \cite{SJE} & 65.6 & 61.9 & 53.9 & 53.7\tabularnewline
ESZSL \cite{ESZSL2015} & 58.2 & 58.6 & 53.9 & 54.5\tabularnewline
SAE \cite{SAE2017} & 53.0 & 54.1 & 33.3 & 40.3\tabularnewline
Sync \cite{SYNC2016} & 54.0 & 46.6 & \textbf{55.6} & 56.3\tabularnewline
\hline 
\textbf{Ours} & \textbf{71.4} & \textbf{65.8} & 52.1 & \textbf{61.7}\tabularnewline
\hline 
\end{tabular}
\caption{Results in the disjoint assumption zero shot setting with the per-class accuracy metric.}
\label{tab2}
\end{table*}

\begin{table*}

\centering{}%
\begin{tabular}{|c|c|c|c|c|}
\hline 
\textbf{Method} & \textbf{AWA-1} & \textbf{AWA-2} & \textbf{CUB} & \textbf{SUN}\tabularnewline
\hline 
DAP \cite{lampert2014attribute} & 0.0 & 0.0 & 3.3 & 7.2\tabularnewline
IAP \cite{lampert2014attribute} & 4.1 & 1.8 & 0.4 & 1.8\tabularnewline
CONSE \cite{CONSE} & 0.8 & 1.0 & 3.1 & 11.6\tabularnewline
SJE \cite{SJE} & 19.6 & 14.4 & 32.8 & 19.8\tabularnewline
ESZSL \cite{ESZSL2015} & 12.1 & 11.0 & 21.0 & 15.8\tabularnewline
Sync \cite{SYNC2016} & 16.2 & 18.0 & 19.8 & 13.4\tabularnewline
DeViSE \cite{DEVISE} & 22.4 & 27.8 & 32.8 & 20.9\tabularnewline
ALE \cite{akata2016label}& 27.5 & 23.9 & \textbf{34.4} & 26.3\tabularnewline
\hline 
\textbf{Ours} & \textbf{47.2} & \textbf{51.2} & \textbf{34.5} & \textbf{26.7}\tabularnewline
\hline 
\end{tabular}
\caption{Results in the Generalized zero shot setting. We use harmonic mean of accuracy on seen and unseen classes as metric.}
\label{tab3}
\end{table*}
\subsection{Features}
Similar to recent papers, we use deep features extracted from Convolutional Neural Networks(CNN) for our experiments. The images from AwA dataset are not publicly available. Hence we use VGG features \cite{VGGNET} provided by the authors for this dataset. Recently \cite{XLSA17} have released a new dataset with the same classes as AwA, however with publicly available images. They call this AwA-2 and the original dataset AwA-1. We use Resnet101 features provided by \cite{XLSA17} for AwA-2, CUB and SUN for fair comparisons. We empirically observe that VGG net features perform slightly better($2\%$) for AwA compared to Resnet-101 features. For Imagenet, we use Alexnet features \cite{AlexNet} for fair comparisons.

The class embedding features also play an equally important role in zero shot learning. For the AwA, SUN, and CUB datasets, we use the attribute annotations provided by the respective authors. However, Imagenet has no such annotated features and we use 1000 dimensional word2vec features \cite{word2vec} trained using a skip-gram model on the Wikipedia corpus, similar to \cite{SSVOC2016}.
\subsection{Train-Test splits: An Important Note}
Most recent zero shot learning models were evaluated on a particular test-train split of classes, or as an average of $n$ (4 or 10) random splits. However as pointed out by \cite{xianCVPR17} there is a significant problem with this approach. Some of the test classes overlap with the training classes of Imagenet on which the feature extraction CNN was pretrained on. This makes the model perform better on such overlapping classes, thereby showing significantly greater accuracy on such classes (contributing to greater overall accuracy). However such an evaluation is not representative of the true performance on the model. Hence \cite{xianCVPR17} provide a novel test train split for each dataset, ensuring that none of the test classes occur in the 1000 classes of Imagenet. They observe that the performance of all methods reduces significantly with such splits. The results obtained from our experiments also corroborate this observation. Thus we present our results on the novel proposed splits of \cite{xianCVPR17}, but  the results on the older splits are also presented for the sake of completeness: on the most widely used AwA dataset, we obtain accuracy of 85.81\% on the standard splits proposed by the authors of the dataset\cite{journals/pami/LampertNH14}. On the fine-grained CUB dataset we 54.3\%. There are no standard splits on the SUN dataset. Hence we experiment with 10 random splits to obtain an average of 88.5\%. However, we would like to reemphasize that these results, although high, are inconclusive in the strictest definition of zero-shot learning.    

\begin{table}
\caption{Results on the Imagenet dataset (top-5 accuracy)}
\begin{centering}
\begin{tabular}{|c|c|}
\hline 
\textbf{Method} & \textbf{Accuracy}\tabularnewline
\hline 
\hline 
AMP\cite{conf/cvpr/FuXKG15} & 13.1\tabularnewline
DeViSE\cite{DEVISE} & 12.8\tabularnewline
CONSE\cite{CONSE} & 15.5\tabularnewline
SS-Voc\cite{SSVOC2016} & 16.8\tabularnewline
\hline 
\textbf{Ours} & \textbf{24.7}\tabularnewline
\hline 
\end{tabular}
\par\end{centering}
\label{tab4}
\end{table}
\subsection{Parameters}
The parameters of the Neural Network are trained with a batch size of 50 and Adam optimizer \cite{ADAM} with a learning rate of $10^{-3},\beta_1=0.9,\beta_2=0.999$. We use glorot initialization \cite{journals/jmlr/GlorotB10} for both the encoder and the decoder. There are two kinds of hyper-parameters in our model. The network hyper-parameters (such as batch size, size of the latent variable) and the SVM cost parameter. The latent variable size was set to 100 and the SVM cost parameter was set to 100 by cross validation on training classes. We empirically observe that the model performance saturates after around 300 generated pseudo data samples.

\subsection{Evaluation Metric}

For AwA, CUB, and SUN we use the average per class accuracy as the metric. Results are compared with those presented in \cite{XLSA17}. Average per class accuracy is a better metric for evaluating imbalanced test sets. It is defined as follows :
\[
acc_{avg}^{per-class}=\frac{1}{|Y|}\sum_{i=0}^{|Y|}\left(\frac{N_{correct}^{(class-i)}}{N_{total}^{(class-i)}}\right)
\]

\cite{XLSA17} observe that due to the class imbalance in the dataset particularly AwA, there is a significant difference (about $4\%$) in the average per class accuracy and the per image: 
\[
acc_{avg}^{per-image}=\frac{N_{correct}}{N_{total}}
\]

For the Imagenet dataset, we measure the top-K accuracy i.e the classification of a test image is correct if the true label occurs in the top K predictions of the model. Similar to \cite{SSVOC2016} and \cite{CONSE}, we set the value of K to 5.

\begin{figure*}
\caption{T-SNE visualization \label{tSNE}}
\begin{centering}
\begin{tabular}{|c|c|}
\hline 
\includegraphics[width=5cm,height=5cm]{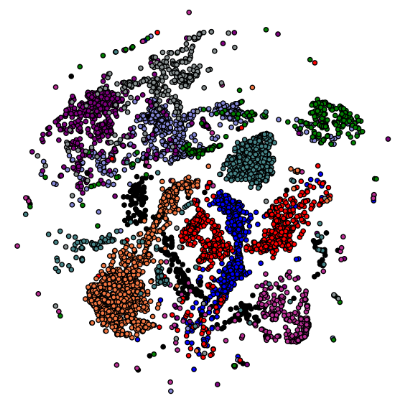} & \includegraphics[width=5cm,height=5cm]{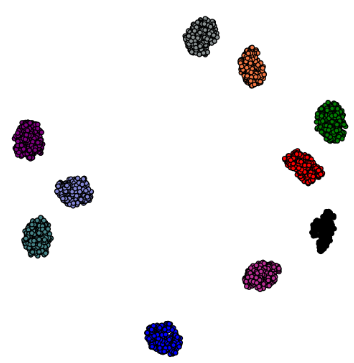}\tabularnewline
\hline 
\end{tabular}
\par\end{centering}

T-SNE Visualization of the true data(left) and the data generated from the network(right) for the unseen classes of AwA dataset (best viewed in color). Note that the data is generated only from the attribute vectors of the class, without looking at even a single image. For most classes, the predicted vectors are close to the true vectors. However, the model suffers from the mode dropping problem (see red, blue).  Recently, many methods for avoiding mode collapse have been studied \cite{journals/corr/RoscaLWM17} which may be applicable to the current problem.
\end{figure*}

\subsection{The Generalized Zero Shot Setting}
For the generalized zero shot setting, we follow the protocol by \cite{GenZSL2016}. From the training images of the seen classes, we set aside $20\%$ of data and train only on the remaining $80\%$. To reduce the bias towards the seen data, the SVM uses the data generated from both the seen and unseen classes as opposed to using the actual data for seen classes. The SVM is evaluated separately on both the set aside seen classes data and the test data from unseen classes. As proposed by \cite{xianCVPR17}, we use the harmonic mean of the two accuracies as a measure of performance for the generalized zero shot setting.
\subsection{Results}
The results on the disjoint assumption zero shot learning with the proposed splits from \cite{xianCVPR17} are presented in Table \ref{tab2}, and for the generalized zero shot in Table \ref{tab3}. 

\subsubsection*{Comparison with previous models } In the disjoint assumption zero-shot setting, the proposed model performs significantly better than state-of-the-art on the coarse-grained datasets AwA-1 and AwA-2. On AwA-1 which is the most widely used benchmark for zero-shot learning, we achieve an improvement of \textbf{5.8\%} in the disjoint assumption. On SUN which is a fine-grained places dataset, the performance gains are significant. On the extremely fine-grained CUB dataset, the performance is slightly less compared to the state-of-the-art. We attribute this to the extremely fine-grained nature of the CUB dataset, where the classes are fairly close. Our intuition is that it is difficult to generate fine-grained image features from just the attributes. 

Imagenet is a much more challenging dataset, particularly due to the lack of explicit attribute vectors. On this we achieve an improvement of about \textbf{7.9\%} over other approaches.  We may note here that currently no other single method claims the best results on all the datasets simultaneously.

In the more realistic generalized zero-shot setting the improvements are even more. In this setting the proposed model improves over state of the art by \textbf{20\%} (absolute) on the coarse grained datasets, while achieving state-of-the-art performance on the fine grained datasets.  We attribute this to two reasons: The underlying distribution is better captured using a Variational Auto-encoder, than by just learning a mapping from image to attribute space and since our model generates image features for seen classes also, it has lesser bias towards the seen classes, which is inherent in other methods. Thus, we may be able to do better on the unseen classes.

We observe that the performance gains are much higher on the harder problem of generalized zero-shot learning. In the disjoint ZSL setting since there are a lesser number of classes, and no confusion between seen and unseen classes, the problem is simpler. Thus, previous approaches may be able to perform well in this scenario. However, in the generalized setting, more sophisticated techniques are necessary, and thus our approach beats competitors by a significant margin.   

\subsubsection*{Visualization}
We visualize the image feature vectors generated by our model for each class using the t-sne \cite{maaten2008visualizing} method in Figure-\ref{tSNE} and compare it with the original test image feature vectors for the AwA-1 dataset. We make the following observations: 1.The generated image features are close to the original feature vectors. This shows that our model is able to capture the underlying image generation process to a good extent which manifests in terms of better performance on the benchmark datasets. 2.The generated images are unimodal. This means that several modes of the underlying distribution are missing from the learned distribution. This offers clues into improving the model to enforce multi-modality learning. This is however, beyond the scope of this work. 

\section{Conclusion}
We have presented a novel approach to zero shot learning by modeling it as a missing data problem. We show that our model compares favorably with the state-of-the-art on four benchmark datasets, while outperforming them for all four datasets in the much harder generalized zero shot setting. 
There are several areas for improvements such as automatically generating attribute vectors for classes using wikipedia articles, end to end training to learn better features for images etc. We leave this for future work. 

{\small
\bibliographystyle{ieee}
\bibliography{egbib}
}

\end{document}